\documentclass[runningheads]{llncs}

\usepackage[OT1]{fontenc}
\usepackage{amsmath,amssymb,amsfonts}
\usepackage{bm}
\usepackage{textcomp}
\usepackage{stmaryrd}

\usepackage{url}
\usepackage{algorithmic}
\usepackage{xcolor}
\usepackage{tikz}

\usepackage{multirow}

\renewcommand{\tabcolsep}{.5em}

\usepackage{listings}

\makeatletter
\AtBeginDocument{%
	\def\doi#1{\url{https://doi.org/#1}}}
\makeatother

\newcommand{\bbb}{\mathbb{B}}
\newcommand{\B}{\mathcal{F}}
\newcommand{\F}{\mathcal{F}}
\newcommand{\K}{\mathcal{K}}
\newcommand{\T}{\mathcal{T}}
\newcommand{\A}{\mathcal{A}}
\newcommand{\Ind}{\mathsf{Ind}(\K)}
\newcommand{\x}{\mathbf{x}}
\newcommand{\X}{\mathbf{X}}
\newcommand{\y}{\mathbf{y}}
\newcommand{\z}{\mathbf{z}}
\newcommand{\Z}{\mathbf{Z}}
\newcommand{\p}{\mathbf{p}}
\newcommand{\Tr}{\mathbf{T}}
\def\cl{\mathrel|\joinrel\sim}
\newcommand{\argmax}{\mathop{\mathrm{argmax}}}
\newcommand{\ber}{\mathrm{Ber}}
\newcommand{\bPi}{\mathbf{\Pi}}
\newcommand{\bmu}{\bm{\mu}}
\newcommand{\bP}{\mathbf{P}}

\begin{document}

\sloppy

\title{Simple and Interpretable\\ Probabilistic Classifiers for Knowledge Graphs}

\author{Christian Riefolo\inst{1,3} \and Nicola Fanizzi\inst{1,2}\orcidID{0000-0001-5319-7933} \and Claudia d'Amato\inst{1,2}\orcidID{0000-0002-3385-987X}}

\institute{LACAM -- Dipartimento di Informatica -- Universit\`a degli Studi di Bari Aldo Moro\\ \email{c.riefolo@studenti.uniba.it}\\ \email{nicola.fanizzi@uniba.it claudia.damato@uniba.it}  \and CILA -- Universit\`a degli Studi di Bari Aldo Moro\and Fincons Group -- \emph{Future Gateway} Delivery Center -- Bari\\ \email{christian.riefolo@finconsgroup.com}}

\maketitle

\begin{abstract}
Tackling the problem of learning probabilistic classifiers from incomplete data in the context of Knowledge Graphs expressed in Description Logics, we describe an inductive approach based on learning simple belief networks.
Specifically, we consider a basic probabilistic model, a Naive Bayes classifier, based on multivariate Bernoullis and its extension to a two-tier network in which this classification model is connected to a lower layer consisting of a mixture of Bernoullis.
We show how such models can be converted into (probabilistic) axioms (or rules) thus ensuring  more interpretability. 
Moreover they may be also initialized exploiting expert knowledge.
We present and discuss the outcomes of an empirical evaluation which aimed at testing the effectiveness of the models on a number of random classification problems with different ontologies.
\end{abstract}


\section{Introduction} 

Classifying individual resources in the context of \emph{Knowledge Graphs} (KGs)~\cite{hogan2021} is a fundamental task enabling several more complex applications.
In this context, classification may turn out to be a non trivial task, even in an approximate form, especially when the original semantics has to be preserved as much as possible.

Current research on graph-structured data focuses on \emph{representation learning} methods~\cite{Wang17-survey,Murphy22pml1} that aim at mapping entities and relations to embedding spaces where tasks like classification and link prediction can be defined in terms of linear algebra operations.
The main downsides with these methods are related both to the difficulty of incorporating implicit knowledge that can be precisely extracted from the KG through deductive reasoning, and to the poor interpretability of complex classification models, as the embedding spaces turn out to be hardly relatable to the original features, so that these models are exploited as black-boxes whose decisions can be hardly explained.

Alternatively, for the sake of interpretability of the models and their decisions, it is possible to resort to simpler yet still effective probabilistic models inferred from data, ultimately defined in terms of  Boolean variables that leverage on basic logic features extracted from primitive classes and properties in the ontology~\cite{Fanizzi22sitis}.
Following some ideas applied in the context of neural learning~\cite{Tresp97mlj}, simple graphical models can be fitted and also converted into probabilistic rules or  simplified to generate  axioms~\cite{Minervini12sac,Fanizzi22sitis} in Description Logics (DL)~\cite{DL-HB-07} ensuring a direct interpretation in terms of the original representation. 

Specifically, we focus on the problem of fitting \emph{multivariate Bernoulli} models on which the common simplifying assumption is made of conditional independence of the input features given the target one, characterizing the Naive Bayes classifiers, which simplifies the model often without compromising its effectiveness~\cite{Domingos97mlj}. 
Then we consider also a two-tier hierarchical model in which a \emph{mixture of Bernoullis}~\cite{Ghahramani93}  is used to cluster the individuals in groups before applying the classifier. 

Generative models are especially suitable for cases in which the data available for their fitting is inherently incomplete, which is very likely when reasoning with an underlying open-world assumption, as typical of Logics and  large (distributed) KGs. Moreover, they may be also used for multiple additional applications, such as KG completion and refinement, axiom (disjointedness) discovery, clustering, anomaly detection, etc.  
Differently from more complex models, these classifiers are also suitable for an easier interpretation, verification and even integration operated by domain experts, or by anybody that shares the intended semantics of the terminology.
For example, in principle existing background knowledge in the form of probabilistic rules or axioms could be integrated within these models. 

An experiment that aimed at testing  feasibility and effectiveness of such models on a number of random classification problems with different Web ontologies is described. 
As a baseline, we also considered a related simple discriminative model like \emph{Logistic Regression}: specifically a regularized version was adopted to enforce the sparseness of the coefficients of the linear combination, so to ensure  model interpretability as much as possible.

In the remainder of the paper, we briefly introduce a Boolean encoding for the individuals in terms of basic features / concepts (Sect.~\ref{sec:basics}). 
Then, in Sect.~\ref{sec:mb}, a simple model based on multivariate Bernoullis is described, together with details on the probabilistic rules/axioms that can be derived, and its fitting, which is further extended in the next Sect.~\ref{sec:EM} discussing better ways to cope with incompleteness, resorting to a procedure based on EM. 
The model can be extended considering a simple two-tier hierarchy as shown in Sect.~\ref{sec:hbm}.
An empirical evaluation is described in Sect.~\ref{sec:exp} with a comparison of some instances of these models on a number of classification tasks on real ontologies. 
Finally Sect.~\ref{sec:end} concludes the paper discussing some limitations and possible extensions of this work.


\section{Preliminaries}\label{sec:basics}

Preliminarily, we assume familiarity with the basic notions and the standard notation of Description Logics as in the following we will focus on knowledge graphs represented through DL axioms and ultimately as OWL-DL ontologies. 

Formally, let $\K$ be a DL \emph{knowledge base} $\K=\langle \T,\A \rangle$, where the \emph{TBox} $\T$ is a set of terminological axioms that regard concepts and roles (i.e.\ classes and properties) of the domain of interest, and the \emph{ABox} $\A$ contains basic assertions, i.e.\ facts regarding the individuals whose collection will be indicated with $\Ind$.


In line with previous works on distances or kernels for these representations~\cite{dAmato08a,Fanizzi12jws}, we will build the models upon a simple encoding of the individuals in a feature space.
In the following the Boolean set will be denoted by $\bbb=\{0,1\}$. 
Given an ordered set $\B=\{ F_i \}_{i=1}^D$ of basic features, i.e.\ concepts in the signature of $\K$ or defined in terms of other concepts and roles (classes and properties) therein\footnote{A set of primitive classes, also including those obtained as restrictions on roles, such as $\exists R.\top$ or $\forall R.\top$. In the experiments we will use those at the bottom of the subsumption hierarchies determined by the axioms in $ \T $.}, we will consider each individual $\texttt{a} \in \Ind$ as represented by the Boolean vector $\mathbf{a}\in \bbb^D$ with each component, $i \in [1:D]$: 
\begin{equation}\label{eq:bool-vec}
	a_i = \begin{cases} 1 & \K \vdash F_i(\texttt{a})\\
		0 & \K \vdash \neg F_i(\texttt{a}) \end{cases}
\end{equation}
where $\vdash$ denotes the underlying proof procedure. 
Note that $a_i$ stands for the Boolean value corresponding to the definite \emph{membership} to $F_i$~($1$) or to its complement~($0$); then for some $F_i$, $a_i$ may be undefined when the procedure $\vdash$ is unable to determine  its membership under the standard open-world semantics adopted in DL. 

These cases may be treated by replacing the missing values with a fixed constant $m_i \in\ ]0,1[ $, i.e. a prior on the uncertain membership to $F_i$, derived from \emph{pseudo-counts} provided by experts or from frequencies observed in the dataset. 
Alternatively, one might consider adopting an uninformative initialization for the missing values in the input vectors and then use a non-supervised model as an encoder (see \cite{Fanizzi22sitis}).

\section{Multivariate Bernoulli Naive Bayes Model}\label{sec:mb}


With individuals represented as $D$-dimensional binary vectors $\x = [x_1,x_2,\ldots,x_D]^T$, and a binary output variable $ y $ indicating the membership w.r.t.\ the target class $ C $, their conditional distribution for either membership case can be modeled as a \emph{Multivariate Bernoulli Naive Bayes Model} (MBNBM), i.e.\ a joint distribution of Bernoulli variables, that are assumed to be conditionally independent given the membership to $C$:
\begin{equation}\label{eq:mb}
	P(\x|y=b) = \ber(\x|\p_b) = \prod_{i=1}^D \ber(x_i|p_{b i})   = \prod_{i=1}^D (p_{b i})^{x_i}(1-p_{b i})^{1-x_i}
\end{equation} 
with parameters $p_{bi} = P(x_i=1|y=b)$, conditional probabilities of $x_i = 1$ (i.e.\ of individual \texttt{x} to belong to  $F_i$) given the membership to $C$ indicated by $ y=b $. 

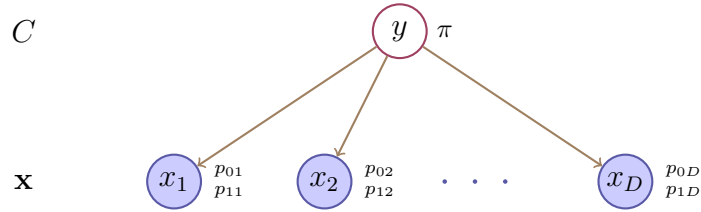
\begin{figure}[tb]
	\centering

\begin{tikzpicture}[draw=gray!50!brown, ->, thick, font=\fontsize{12}{0}, 
]

\tikzstyle{eti}=[font=\fontsize{10}{0}]
\tikzstyle{nodo}=[circle, draw=blue!30!gray, fill=blue!20!white, inner sep=0pt, minimum size=2.2em]

\node (x) at (0,0) {$\mathbf{x}$};
\node[nodo, label={[eti]0:$ p_{01} \atop p_{11} $}] (x1) at (2,0) {$x_1$};
\node[nodo, label={[eti]0:$ p_{02} \atop p_{12} $}] (x2) at (4,0) {$x_2$};
\node[color=blue!30!gray] (ell1) at (6,0) {\huge . . .};
\node[nodo, label={[eti]0:$ p_{0D} \atop p_{1D} $}] (xM) at (8,0) {$x_D$};

\node (C) at (0,2) {$C$};
\node[nodo, label={[eti]0:$ \pi $}, draw=purple!50!gray, fill=white] (y) at (5,2) {$ y $}; 

\draw (y) -- (x1);
\draw (y) -- (x2);
\draw (y) -- (xM);

\end{tikzpicture}
	\caption{Multivariate Bernoulli Naive Bayes Model as a belief network}\label{fig:mb}
\end{figure}

The resulting joint distribution $P(\x,y) = P(\x|y)P(y)$ is depicted in Fig.~\ref{fig:mb} as a (Naive Bayes) belief network, where together with the $p_{bi}$'s, another parameter $ \pi $ is required for specifying the \emph{prior}\footnote{This setting naturally extends to one with multiple mutually disjoint target classes, say $ C_1,\ldots,C_M $, in which $ y $ can assume one value from $\{1,\ldots,M\}$ indicating one of the target classes.} of the output $y$, i.e.\ $ P(y=1) = \pi_1 = \pi$, hence $ P(y=0) = \pi_0 = 1 - \pi$.

\subsection{Classification}

A classification problem amounts to estimating the state of the output $ y $ for an input individual \texttt{x} represented by the Boolean vector $\x$. 
Given the model parameters, i.e.\ prior  $P(y)=\pi$ and $ P(\x|y)$, the posterior can be computed using Bayes' rule and Eq.~\eqref{eq:mb}:
\begin{eqnarray}\label{eq:classifier-mb}
	P(C|\x) &=& P(y=1|\x) = \frac{P(\x|y=1)P(y=1)}{P(\x)}\nonumber \\ &=&  \frac{\pi \ber(\x|\p_1)}{\pi \ber(\x|\p_1) + (1-\pi) \ber(\x|\p_0)}
\end{eqnarray}
$P(\neg C|\x) = 1-P(C|\x)$, then the membership to be predicted depends on the more probable case:
\[
\hat{y} = \argmax_{b \in \bbb} P(y=b | \x) 
\]
which is equivalent to the decision procedure: 

\begin{equation}\label{eq:decision}
	\textrm{if}\ P(y=1|\x) > 0.5\ \textrm{then}\ \K \cl C(\texttt{x})\ \textrm{else}\ \K \cl \neg C(\texttt{x})
\end{equation}
where a different symbol $\cl$ is used to denote the prediction based on the probabilistic model (instead of the logical one $ \vdash $).
The mid-point $0.5$ is adopted for simplicity, yet more complex decision procedures can be devised, taking into account the case of \emph{rejection} when the probability is close to this value. 
This can be defined in terms of a cost-sensitive decision using a threshold $ \theta  \in (0.5,1] $ for either decision.

\subsection{Rules and Conjunctive Class Definitions}

As discussed in \cite{Fanizzi22sitis}, in line with similar approaches for models based on features with continuous distributions (e.g.\ see~\cite{Ghahramani93,Tresp97mlj}), 
a probabilistic \emph{conjunctive rule} for $C$ may be defined as shown in Fig.~\ref{fig:rule}, which can be simplified considering for each $ i\in [1:D] $ the conjunct with higher probability.
Rules for the prediction of the complement $\neg C$ can be formed analogously, in terms of the parameters $p_{0 i}$ and $ \pi_0 $.

\begin{figure}[bt]
\begin{lstlisting}
	IF $\K \cl C(\texttt{x})$ with prior $ \pi$ THEN each $x_i$ is independently Bernoulli distributed
		AND $ x_1 = 1$ with probab. $p_{11}$ AND $x_1=0$ with probab. $1-p_{11}$ 
		AND $ x_2 = 1$ with probab. $p_{12}$ AND $x_2=0$ with probab. $1-p_{12}$
	  	   (*@\raisebox{-1pt}[0pt][0pt]{$\vdots$}@*)
		AND $ x_D = 1$ with probab. $p_{1D}$ AND $x_D=0$ with probab. $1-p_{1D}$
\end{lstlisting}
\caption{Rule schema}\label{fig:rule}
\end{figure}

Given a minimal threshold $\theta$, as hinted above, an approximate logic definition for the target class may be extracted from a MBNB model considering the sets $ \F^+ = \{i \in [1:D] \mid p_{1i} > \theta\}$ of features  positively correlated with the membership and $ \F^- = \{i \in [1:D] \mid p_{0i} > \theta \}$ of features negatively correlated with the membership. 
Then it is possible to define the approximate axiom:
$$C \sqsubseteq \bigsqcap_{i\in \F^+} F_i\ \sqcap\ \bigsqcap_{j \in \F^-} \neg F_j $$
which can be proposed to a domain expert for its validation and possible inclusion in the definition of the target class $C$.
Note that only some of the basic features are considered, namely those strongly positively/negatively correlated with the membership, where the strength is quantified by $\theta$.

\subsection{Fitting the Model} 

We assume the availability of a complete \emph{training set} $\Tr = \langle \X, \y \rangle = \{(\x^t, y^t)\}^N_{t=1} $ where $\x^t$ is the encoding of an individual $ \texttt{x}^t \in \Ind$ and $ y^t \in \bbb$ indicates the actual membership to $C$.

The naive Bayes classifier can be trained by finding the maximum likelihood (or even the maximum a posteriori) estimate for the parameters $ \bPi =\{\pi, \p_1,\p_0\} $. 
As the probability for a single example is given by \[ 
P(\x^t,y^t|\bPi) = P(y^t|\pi_{y^t}) P(\x^t|y^t, \p_{y^t}) = \pi_{y^t} \ber(\x^t|\p_{y^t})
\]
then the \emph{log-likelihood} $ \mathcal{L}(\bPi) = \log P(\Tr |\bPi) $ can be written:
\begin{align*}
	\mathcal{L}(\Tr|\bPi)  & = \log\prod_{t=1}^N P(\x^t,y^t|\bPi) = \sum_{b\in\bbb} N_b\log\pi_b + \sum_{b\in\bbb}\sum_{t:y^t=b} \sum_{i=1}^D \log \ber(x_i^t|p_{bi}) 
\end{align*}
where $N_b = \sum_t 1(y^t = b)$ are the counts of training examples for either membership case ($b\in \bbb$).

The MLE for the proportions of the prior $ P(y=b) $ can be estimated as $\hat{\pi}_b = {N_b}/{N}$.
As all input features are Bernoulli for either value of $y$, i.e.\ $x_i~|~y~\sim~\ber(p_{yi})$, the MLE for the parameters are $\hat{p}_{bi} = {N_{b i}}/{N_b},\ i \in [1:D]$ with $ N_{b i} = \sum_t 1(x^t_i=1, y^t=b) $.

Actually, one does not expect the features to be independent (conditionally on $ y $). 
However, as discussed in~\cite{Domingos97mlj}, a NB model can still be quite effective, even when this assumption does not hold true, given the limited number of parameters which makes it less prone to overfitting.
To better avoid this problem, one may resort to a Bayesian approach considering Beta conjugate priors~\cite{Murphy22pml1}.

\section{Dealing with Incomplete Data}
\label{sec:EM}

In the context of DL knowledge graphs it is possible that given a \emph{feature} concept the membership of an individual cannot be ascertained logically because of the open-world semantics.
In such non unlikely cases, some input feature $x^t_i$ or target feature $y^t$ would assume an indefinite value making $\Tr = \langle \X, \y  \rangle$ incomplete.

A further assumption should be made about the cause  of missing values for target and/or input features~\cite{LitteRubin19}. 
A MAR (\textit{Missing at Random}) setting is likely not adherent to the ground truth about the data but it is adopted to keep the model simple. 

Although in principle the problem of determining the indefinite values of features could be treated homogeneously over $\Tr$ for missing values either in $\X$ or in $\y$, we  consider distinct cases tackled in subsequent phases: 
\begin{enumerate}
	\item examples with an indefinite value for some input feature $x^t_i$;
	\item examples with an indefinite value for the target feature $y^t$. 
\end{enumerate}

\subsection{Phase 1: Missing Input Values}
\label{sec:ph1}
A simple solution to the problem of missing values of input features is \emph{imputing} a fixed value or a constant decided on a per feature basis (e.g.\ estimated as the mode or the mean of the column vector $\X^T_i$) which may be also conditioned to a known value of the corresponding target feature $y^t$, as previously treated in the estimation of the parameters.

Assuming that the \textit{missingness} of a feature is not informative about its potential value, if parts of an input vector $ \x $ were missing during training (and/or testing), then one may marginalize out such features~\cite{Murphy22pml1}. 
Writing $\x$ as $(\x_o,\x_u)$, with $\x_u$ sub-vector defined on the features with \emph{unknown} (missing) values and $\x_o$ defined on those with \emph{observed} values, classification would amount to maximize: 
\begin{align*}
	P(y|\x_o, \bPi) \propto\ & \pi_y P(\x_o | y, \p_y) = \pi_y \sum_{x_u \in \x_u}\sum_{x_u \in \bbb} P(\x_o, x_u | y, \p_y) =  \pi_y  \ber(\x_o | y, \p_y) 
\end{align*}
with the term for $ \x_u $ ignored because of the assumed conditional independence~\cite{Murphy22pml1}.
More exactly:
\begin{eqnarray*}
	P(y=b | \x_o ) &=& \dfrac{P(\x_o|y=b)P(y=b)}{\sum_{b'\in\bbb} P(\x_o|y=b')P(y=b')} = \dfrac{\ber(\x_o|\p_b)\pi_b}{\sum_{b'\in\bbb} \ber(\x_o|\p_{b'})\pi_{b'}}\\ &=&
	\dfrac{\pi_b\prod_{x_o \in \x_o}\ber(x_o|p_{bo})}{\sum_{b'\in\bbb} \pi_{b'}\prod_{x_o \in \x_o}\ber(x_o|p_{b'o})}
\end{eqnarray*}

To estimate the value of any $x_u$ based on the observed features $\x_o$, one can compute:
\[
P(x_u|\x_o) = \dfrac{\sum_b \pi_b \ber(x_u|\p_b) \ber(\x_o|\p_b) }{\sum_b \pi_b \ber(\x_o|\p_b)}
\]
Then the expected value would be: 
\begin{equation}
	\mathbb{E}[x_u|\x_o] = \sum_{v\in \bbb} v P(x_u=v|\x_o) =  \dfrac{\sum_b w_b \pi_b \ber(\x_o|\p_b) }{\sum_b \pi_b \ber(\x_o|\p_b)}
\end{equation}
where $w_b = p_{bu}$.


Another option (which is valid even when the naive Bayes assumption cannot be not made) is to fit the model in case of incomplete data resorting to an \emph{Expectation-Maximization} (EM) procedure~\cite{Ghahramani93,Murphy22pml1}. 
The goal is to find maximum likelihood solutions for the latent variables $\Z$ (in this case the $\x^t_u$ in $\Tr$) to be estimated together with the model parameters~$\bPi$. 
On each iteration $l$, in the \emph{E~step}, the current parameter values $\bPi^{l}$ are used to find the distribution of the latent variables conditional on the observed ones (including the current values for $\bPi^{l}$). 
Then, using this posterior, the \emph{expectation of the complete-data log likelihood} is computed: $\mathcal{Q}(\bPi | \bPi^{l}) \gets \mathbb{E}[\mathcal{L}(\bPi | \Z, \Tr) | \Z, \Tr, \bPi^{l}] $, for some general parameter value $ \bPi $. 
In the \emph{M~step}, the revised parameter estimate $ \bPi^{l+1} $ is determined by maximizing $\mathcal{Q}(\bPi | \bPi^{l})$. 
Before ending each iteration the current parameters are updated for the next step.


In this first phase:

\begin{itemize}
	
\item\textsf{E step} $\forall t\in[1:n], \x^t=\langle \x_o,\x_u \rangle$: \[\ \hat{P}(\x_u =b | \x_o, \bPi^{l}) \gets \sum_{b\in \bbb} \dfrac{\hat{\pi}_b \ber(\x_o|\hat{\p}_{b})}{\sum_{b'} \hat{\pi}_{b'} \ber(\x_o|\hat{\p}_{b'})}\]

\item\textsf{M step} $\forall b\in \bbb, i \in [1:D]$:
\begin{align*}
\hat{p}_{bi} \gets \dfrac{\sum_t \hat{P}(\x_u =b | \x_o, \bPi^{l}) x^t_i}{\sum_t \hat{P}(\x_u =b | \x_o, \bPi^{l})}\qquad \hat{\pi}_b \gets \frac{1}{N} \sum_t \hat{P}(\x_u = b | \x_o, \bPi^{l}) 
\end{align*}
\end{itemize}

\subsection{Phase 2: Missing Target Values} 
As regards the second case, considering a training set $\Tr'$ with a completed $\X$ and a possibly incomplete $\y$, the log-likelihood to be maximized would be:
\[\mathcal{L}(\bPi|\Tr') = \sum_{t=1}^N \log \left[\sum_{b\in \bbb} \pi_{b} \ber(\x^t|\p_{b}) \right] \]
with the known problem of the sum under the logarithm. 
This is another case where the incomplete $\y$ and the model parameters, initialized with the values found in Phase~1,  can be fitted by an EM procedure with the steps 
defined as follows:

\begin{itemize}
	\item \textsf{E step}: $\forall t\in[1:n]$: 
\[
\hat{P}(y^t=b|\x^t) \gets \dfrac{\hat{\pi}_b \ber(\x^t|\hat{\p}_{b})}{\sum_{b'} \hat{\pi}_{b'} \ber(\x^t|\hat{\p}_{b'})} 
\]

\item \textsf{M step}: $\forall b\in \bbb, i \in [1:D]$:\\
\[
\hat{p}_{bi} \gets \frac{\sum_t \hat{P}(y^t=b|\x^t) x^t_i}{\sum_t \hat{P}(y^t=b|\x^t)}\qquad 
\hat{\pi}_b \gets \frac{1}{N} \sum_t \hat{P}(y^t=b|\x^t) 
\] 
\end{itemize}



\section{A Hierarchical Bernoulli Model}\label{sec:hbm}

In general, the features for a class are likely to be correlated then a single model for describing  the instances'  distribution may turn out to be inadequate when the assumption of conditional independence cannot be made, leading to an inaccurate classifier. 

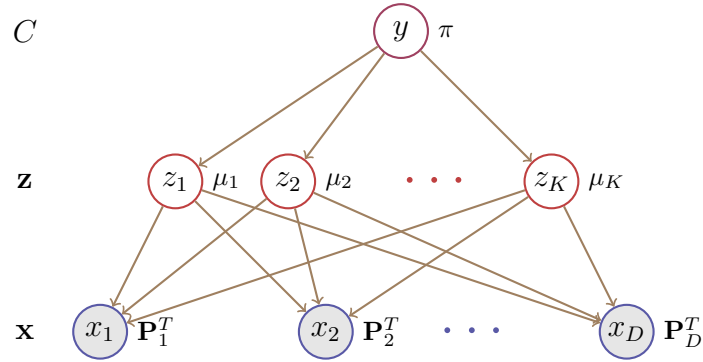
\begin{figure}[tb]
	\centering\begin{tikzpicture}[draw=gray!50!brown, ->, thick, font=\fontsize{12}{0}, ]

\tikzstyle{eti}=[font=\fontsize{10}{0}]
\tikzstyle{nodo}=[circle, draw=blue!30!gray, fill=white, inner sep=0pt, minimum size=2.2em]

\node (x) at (0,0) {$\mathbf{x}$};
\node[nodo, fill=gray!20, label={[eti]-0:$ \bP^T_1 $}] (x1) at (1,0) {$x_1$};
\node[nodo, fill=gray!20, label={[eti]-0:$ \bP^T_2 $}] (x2) at (4,0) {$x_2$};
\node[color=blue!30!gray] (ell1)                          at (6,0) {\huge $\cdots$};
\node[nodo, fill=gray!20, label={[eti]-0:$ \bP^T_D $}] (xD) at (8,0) {$x_D$};

\node (z) at (0,2) {$\mathbf{z}$};
\node[nodo,draw=red!50!gray, label={[eti]0:$ \mu_1 $}] (z1) at (2,2) {$z_1$};
\node[nodo,draw=red!50!gray, label={[eti]0:$ \mu_2 $}] (z2) at (3.5,2) {$z_2$};
\node[color=red!50!gray] (ell2)                              at (5.5,2) {\huge $\cdots$};
\node[nodo,draw=red!50!gray, label={[eti]0:$ \mu_K $}] (zK) at (7,2) {$z_K$};

\node (C) at (0,4) {$C$};
\node[nodo,draw=purple!50!gray,fill=,label={[eti]0:$ \pi $}] (y) at (5,4) {$y$};

\draw (z1) -- (x1);
\draw (z1) -- (x2);
\draw (z1) -- (xD);

\draw (z2) -- (x1);
\draw (z2) -- (x2);
\draw (z2) -- (xD);

\draw (zK) -- (x1);
\draw (zK) -- (x2);
\draw (zK) -- (xD);

\draw (y) -- (z1);
\draw (y) -- (z2);
\draw (y) -- (zK);

\end{tikzpicture}
	\caption{Hierarchical Bernoulli Model}
	\label{fig:mmb}
\end{figure}

This problem can be tackled by considering a distribution modeled as a mixture of (multivariate) Bernoullis.
Then, introducing an intermediate $ K $-dimensional layer $ \mathbf{z} $ of latent binary indicator variables corresponding to the nodes $\{z_k\}_{k\in[1:K]} $, characterized by a different Bernoulli $ P_k(\x|z_k) $, a 2-tier model is defined whose structure is depicted in Fig.~\ref{fig:mmb}.
This network, that will be referred to as \emph{Hierarchical Bernoulli Model} (HBM), combines the mixture of multivariate Bernoullis (bottom-tier) with the former MBNB classifier (top-tier).
Alternatively, this model can be also described in the form of a \emph{Hierarchical Mixture of Experts}~\cite{Jordan93}.

In this case, given the model parameters $ \bPi = \{\pi, \bmu, \bP \} $, where $\mu_k = P(z_k~=~1)$, $\bP = [\p_{k}] \in [0,1]^{K \times D}$ with $\p_k = [p_{k 1},\ldots,p_{k D}]$ (in Fig.~\ref{fig:mmb} we have a column $\bP^T_i$ at each bottom node), we can write: 

\begin{align}
	P(\x|\bPi) &= \sum_{k=1}^K P(\x, z_k=1|\bPi) =  \sum_{k} P(z_k=1|\bmu) P_k(\x|z_k=1; \p_k)  \nonumber \\
	& = 
	 \sum_{k} \mu_{k} \ber(\x | \p_k) \label{eq:hbm}
\end{align} 

The structure depends on the dimensionality of the mixture $ K $, whose choice can be made in advance exploiting part of the training data, to maximize a score function like the BIC~\cite{Murphy22pml1}, for example.
Alternatively a Bayesian procedure can be adopted.

\subsection{Classification}

To decide the membership to $C$ for an input individual \texttt{x} with corresponding Boolean vector $\x$, using Eq.~\eqref{eq:hbm}, the posterior is determined as:  
\begin{eqnarray}\label{eq:classifier-mmb}
	P(y=1|\x; \bPi) &=&  \frac{P(\x|y=1; \bPi) P(y=1)}{P(\x | \bPi)} \nonumber \\  &=&   \frac{\pi \sum_{k=1}^K \mu_{k}  \ber(\x | \p_{k})}{\pi \sum_{k=1}^K \mu_{k}  \ber(\x | \p_{k}) + (1-\pi) \sum_{k=1}^K \mu_{k} \ber(\x | \p_{k})}
\end{eqnarray}

The decision procedure is analogous to (\ref{eq:decision}).
Alternatively we could use disjunctive and conjunctive classification rules translating the hierarchical model.

\subsection{Disjunctive Definitions / Rules}
\label{sec:hbmm-rules} 

Given a MMBM, an axiom for class $C$ may be extracted by defining the class in terms of the components: 
$$C \sqsubseteq \bigsqcup_{k=1,\ldots,K:\atop z_k=1} C_k$$
which can be specified probabilistically indicating the class prior $ \pi $ and $ \mu_{k} $ such that $z_k=1$
or, alternatively, also as a \emph{disjunctive classification rule} for $C$ and a set of probabilistic rules for the $C_k$'s as shown in Fig.~\ref{fig:rule2}. 

\begin{figure}[bt]
\begin{lstlisting}
	IF $\K \cl C(\texttt{x})$ (with prior $\pi$) THEN 
		$\forall k\in [1:K]\!: z_k = 1$ with probab. $\bmu_k$

	IF $\K \cl C_k(\texttt{x})$ THEN  
		AND $ x_1 = 1$ with probab. $p_{1 k}$ AND $x_1=0$ with probab. $1 - p_{1 k}$ 
		AND $ x_2 = 1$ with probab. $p_{2 l}$ AND $x_2=0$ with probab. $1 - p_{2 k}$   
		...
		AND $ x_D = 1$ with probab. $p_{D k}$ AND $x_D=0$ with probab. $1 - p_{D k}$ 
\end{lstlisting}
\caption{Rules from the Hierarchical Model}\label{fig:rule2}	
\end{figure}

For the sake of interpretability, these rules may be simplified considering, for each $ i $, only the more likely sub-case with a minimal threshold to be exceeded. 
A class definition can be given also for the complement $ \neg C $ in terms of the sub-classes $ C_k $ corresponding to the cases when $ z_k=0$.

\subsection{Fitting the Model} 

The network parameters can be determined using a \emph{training set} $\Tr = \langle \X, \y  \rangle$. 
The two tiers can be trained separately: the bottom one tackles an unsupervised learning task, while the top tier is essentially a MBNB classifier. 

As regards the bottom-tier of the structure, the parameters for the mixture model can be learned through a EM~procedure with the cases of incomplete input data treated similarly to the Phase~1, described in Sect.~\ref{sec:EM}.

In this case, given the model parameters $ \bPi $, the complete log-likelihood to be maximized is:
\begin{eqnarray}
	\mathcal{L}(\bPi,\Z|\X)  =  \log \prod_{t=1}^N P(\x^t | \bPi) = \sum_{t=1}^N \log \left[ \sum_{k=1}^K \mu_k  \ber_D(\x^t|\p_k) \right] 
\end{eqnarray}
Note that it also targets the latent variables $ \z$.

and the EM steps can be defined as follows:
\begin{itemize}
	\item \textsf{E-step}: 
	the responsibility of each component of the mixture is estimated as:
	\begin{align*}\label{eq.e-step2}
		\hat{P}(z_k^t|\x^t) \gets  \frac{\hat{\mu}_{k}  P_k(\x^t| \hat{\p}_{k})}{\sum_{h=1}^K \hat{\mu}_{h} P_h(\x^t| \hat{\p}_{h})};
	\end{align*}
	\item \textsf{M-step}: the parameters values are updated on the ground of the new estimates for $ \hat{P}(z^t_k|\x^t) $ along the following rules:
	\begin{eqnarray*}\label{eq.m-step2}
	\hat{\p}_{k} \gets \frac{\sum_{t=1}^N \hat{P}(z_k^t|\x^t) \x^t}{\sum_{t=1}^N \hat{P}(z_k^t|\x^t)}; \qquad 
	\hat{\mu}_{k} \gets  \frac{1}{N} \sum_{t=1}^{N} \frac{\hat{P}(z_k^t | \x^t)}{\sum_{h=1}^K \hat{P}(z_{h}^t |\x^t)}.  
	\end{eqnarray*}
\end{itemize}

Alternatively one may adopt a more sophisticate solution based on the Bayesian treatment of the parameters (using variational inference or Dirichlet processes~\cite{Murphy22pml1}).

The top-tier MBNBM is trained as described in Sect.~\ref{sec:EM}.

\section{Experiments} 
\label{sec:exp}

We briefly present how the models involved in the experiments have been implemented, then we describe the experimental setup (datasets, problems, hyper-parameters setup) and finally we discuss the results of this empirical evaluation. 

\subsection{Implementation}
The implementation of the models' prototypes combines various facilities provided by Python libraries.
\texttt{Owlready2}\footnote{\url{https://pypi.org/project/Owlready2/}} is intended for ontology-oriented programming and  was used to manage the knowledge graphs, including loading and storage while it relies on an embedded (Java) version  of \texttt{Pellet} for reasoning services, mainly used to initialize the binary encoding of the individuals in each ontology.
Libraries \texttt{Scikit-Learn}\footnote{\url{https://pypi.org/project/scikit-learn/}} and \texttt{Sklearn-Bayes}\footnote{\url{https://github.com/AmazaspShumik/sklearn-bayes}} offer various implemented models that could be extended and facilities for the organization of the experiments (measures, cross-validation, etc.) that were employed as described in the following. 
In particular, the HBM was implemented by pipelining a variational Bayes variant of the Bernoulli mixture model with a MBNB classifier.

\subsection{Settings}

Three simple probabilistic models  were assessed in the experiment, namely a basic implementation of the MBNBM, its integration with EM procedures for fitting latent variables and parameters, and  regularized \emph{Logistic Regression}, with a L\textsubscript{1} penalization, as a baseline discriminative probabilistic model of comparable complexity in the number of parameters. 

The KGs of four ontologies were selected to create  problems and related datasets employed as a testbed: \textsc{KRKZeroOne} (a small ontology derived from a well-known UCI dataset), \textsc{New Testament Names} (\textsc{NTNames}), \textsc{Financial}, and an ontology that was generated using the \textsc{Lehigh University Benchmark} (\textsc{LUBM}). 
Ontologies, target classes, code and output files of the experiments are publicly available in the project repository\footnote{\url{https://github.com/THESCREAMINGMONKEY/MBM-EM}}. 

For each ontology basic features (minimal classes in the subsumption hierarchy) have been extracted, then classes formed as universal and existential restrictions were also included, each involving one the available object properties.
However, to define the final set $\F$ and the encoding of the individuals for each dataset, a preliminary univariate \emph{feature selection} phase was performed to pick the most informative ones, i.e.\ those that exhibited a higher variance, with respect to a cutoff threshold.  
Tab.~\ref{tab:numbers} reports the numbers of classes and object properties selected per KG as well as the numbers of (generated/selected) features that were considered in the experiments.

\begin{table}[tb]
	\caption{Numbers of defined classes and object properties per knowledge graph, generated and selected features, individuals}\label{tab:numbers}
	\centering 
	\begin{tabular}{@{ }rccccc@{ }}
		\hline
		\textbf{KGs} & \#classes & \#o.prop's & \#generated & \#selected  & \#ind's \\
		\hline
		\textsc{KRKZeroOne}
		& 7  & 37 & 44 & 28 & 420 \\
		
		\textsc{NTNames}
		& 47  & 27 & 82 & 19 & 676 \\ 
		
		\textsc{Financial}
		& 59  & 16 & 75 & 13 & 1000 \\
		
		\textsc{LUBM}
		& 43  & 25 & 75 & 13 & 1156 \\
		
		\hline
	\end{tabular} 
\end{table}

Artificial learning problems have been randomly created defining 10 new target (disjunctive) classes for each ontology, based on the respective signature. 
To avoid trivial problems, each eligible target class was required to partition the set of individuals in nonempty subsets with a minimal number of positive and negative instances: the ground truth, i.e.\ the reference classification of all individuals in the ontology w.r.t.\ each problem\footnote{Note that all problems were run independently, hence all individuals were exploited as examples in each experiment, with the class-label depending on the target class.} (positive, negative or unlabeled) was determined using the reasoner and encoded in the respective target columns.
It is worthwhile to note that most of the problems were imbalanced and the dataset featured large proportions of unlabeled individuals.

To train the basic MBNB model, missing values in the input encoding have been treated by assuming an uninformative initialization for the likelihood of membership to the various basic $ F_i $'s.
With MBNBM-EM and HBM, the missing values for the input features instances were preliminarily imputed via EM (as described in Subsect.~\ref{sec:ph1}).

For each problem, a \emph{10-fold cross validation} was performed randomly splitting the examples (positive, negative and unlabeled individuals w.r.t.\ the target class) in a \emph{stratified} fashion to preserve the proportions in each fold. 
The metrics that were measured are \emph{precision}, \emph{recall},  \emph{F\textsubscript{1}-measure}, and \emph{geometric mean score} on binary problems, i.e.\ considering only test examples labeled with a definite membership.
Note that, as most of the problems featured imbalanced targets, the average measures were  weighted along the respective proportions of  examples.

The values for the hyperparameter of the Bernoulli mixture (lower tier of the HBM), namely the \textit{number of components} has been tuned beforehand through grid-search in the range $[2,3,\ldots,10]$ on a per-ontology basis. 
A series of further implementation-specific parameters is available (see the  \texttt{Sklearn-Bayes} library documentation). 
We adopted a conservative setup for them, e.g.\ limiting the \textit{number of random restarts} to 10, as a trade-off between model effectiveness and training efficiency. 
A detailed list of these settings can be found in the project repository.

\subsection{Results}

The outcomes of the tests in terms of the measures averaged over the various problems per KG are summarized in Tab.~\ref{tab:results}.  
More details on each experiment can be found in the output files made available.

\begin{table}[tb]
	\caption{Outcomes of the experiments with the models:  \emph{precision}~(P), \emph{recall}~(R),  \emph{F\textsubscript{1}-measure}~(F\textsubscript{1}), and \emph{Geometric mean score}~(G\textsubscript{m}) ± standard deviation averaged over the 10 classification problems per KG}
	\label{tab:results}\centering
	\begin{tabular}{@{}r@{\; }c@{\; }c@{\; }c@{\; }c@{\; }l@{}}
		\hline
		\multirow{2}{*}{\textbf{Ontologies}/\textbf{KGs}} &   \multicolumn{4}{c}{\textbf{models}} & \\
		\cline{2-5}
		& MBNB & MBNB-EM & HBM & \textsc{LogReg} \\
		\hline
		\multirow{4}{*}{\textsc{KRKZeroOne}}
		& 0.995 ± 0.002  & 0.999 ± 0.002 & 0.999 ± 0.002 & 0.943 ± 0.092 &  P \\
		& 0.995 ± 0.003  & 0.999 ± 0.002 & 0.999 ± 0.003 & 0.148 ± 0.243 &  R \\
		& 0.995 ± 0.003  & 0.999 ± 0.002 & 0.999 ± 0.003 & 0.113 ± 0.187 &  F\textsubscript{1}\\
		& 0.995 ± 0.003  & 0.999 ± 0.002 & 0.995 ± 0.008 & 0.146 ± 0.205 &  G\textsubscript{m}\\
		\hline
		\multirow{4}{*}{\textsc{NTNames}} 
		& 0.976 ± 0.024  & 0.981 ± 0.019 & 0.963 ± 0.022 & 0.941 ± 0.074 &  P\\
		& 0.950 ± 0.060  & 0.978 ± 0.020 & 0.957 ± 0.030 & 0.524 ± 0.436 &  R\\
		& 0.958 ± 0.045  & 0.973 ± 0.028 & 0.946 ± 0.035 & 0.491 ± 0.417 &  F\textsubscript{1} \\
		& 0.876 ± 0.195  & 0.777 ± 0.301 & 0.552 ± 0.369 & 0.173 ± 0.172 &  G\textsubscript{m} \\  
		\hline
		\multirow{4}{*}{\textsc{Financial}} 
		& 0.993 ± 0.009 & 0.993 ± 0.009 & 0.937 ± 0.056 & 0.900 ± 0.067 & P\\
		& 0.975 ± 0.041 & 0.981 ± 0.022 & 0.924 ± 0.073 & 0.873 ± 0.094 & R\\
		& 0.981 ± 0.029 & 0.983 ± 0.025 & 0.896 ± 0.100 & 0.822 ± 0.129 & F\textsubscript{1} \\
		& 0.891 ± 0.230 & 0.836 ± 0.274 & 0.464 ± 0.308	& 0.335 ± 0.188 & G\textsubscript{m} \\  
            \hline
		\multirow{4}{*}{\textsc{LUBM}}
		& 1.000 ± 0.000 & 1.000 ± 0.000 & 0.969 ± 0.038 & 1.000 ± 0.000 & P\\
		& 1.000 ± 0.000 & 1.000 ± 0.000 & 0.962 ± 0.047 & 0.656 ± 0.004 & R \\
		& 1.000 ± 0.000 & 1.000 ± 0.000 & 0.946 ± 0.067 & 0.672 ± 0.002 & F\textsubscript{1}\\
		& 1.000 ± 0.000 & 1.000 ± 0.000 & 0.874 ± 0.059	& 0.702 ± 0.357 & G\textsubscript{m} \\  
		\hline
	\end{tabular}
\end{table}

Considering the various measures, the general trend of the outcomes shows that MBNB-EM achieved the best performance w.r.t.\ all metrics and ontologies, with MBNB as a close runner-up, and both outperformed \textsc{LogReg}.
The outcomes show that the performance tends to increase with larger datasets in terms of numbers of individuals: they likely make it easier to get better estimates for parameters and latent variables.
Also the hierarchical model proved its effectiveness although it was not as effective as the simpler models over all problems hinting that larger datasets would be required to properly fit a model with more parameters. Moreover a fine-tuning of the hyperparameters on a per-problem basis (target class) would produce better models at the cost of an extra computational effort. 

The lesser performance of \textsc{LogReg} was likely also due to the amount of penalty adopted to produce simpler models (i.e.\ with few non-zero parameters).
Separate experiments showed that its performance could be improved adopting a L\textsubscript{2} penalization (and fine-tuning the amount of regularization with each dataset). However, the resulting models would turn out to be  less interpretable.

As regards the stability of the results, it must be pointed out that the outcomes for each ontology were averaged over the 10 problems. 
Overall, it appears that the proposed MBNB models are also more stable than the considered baseline.
The diversity of the target class determined some limited variance especially in terms of recall and consequently of F\textsubscript{1}-measure.
For a deeper insight in the outcomes of the single learning problems the output files can be consulted.

A \emph{Friedman-Nemenyi} test was performed  which revealed the experimental results to be statistically significant especially in the comparison of the MBNB models with the baseline.
More specifically, the differences were found significant in the comparison of MBNB with \textsc{LogReg} on \textsc{NTNames} and \textsc{Financial} and of MBNB-EM with \textsc{LogReg} on \textsc{KRKZeroOne}, \textsc{NTNames}, and \textsc{Financial}. 
Further details can be found in the material available in the project repository.

As a final general consideration regarding the interpretability of the models learned, we recall that the basic features involved are selected among those extracted per knowledge graph (see Tab.~\ref{tab:numbers}).
They correspond to the nodes at the input level of the related networks. 
Hence the number of boolean features that can be set in a model (i.e.\ those whose probability may exceed the threshold) is bounded above by the number of these features, $ D $.

\section{Conclusions and Possible Extensions}
\label{sec:end}

In order to better tackle the inherent incompleteness of the DL knowledge graphs, we extended the methods to generate simple probabilistic classifiers leveraging on the semantics of basic logic features regarded as discrete random variables. 
These models are especially suitable for problems related to incomplete data, which is the case of the DL KGs. Besides, they can be converted into probabilistic rules or axioms  with a straightforward  interpretation in terms of the terminology employed by the KG, hence enforcing the model interpretability.
An experiment that tested the effectiveness of these models in comparison with a simple discriminative one (logistic regression) adopted as a baseline, proved them effective and worth of further investigations.


A number of limitations would require our attention.
KGs containing plenty of classes and properties may offer numerous features for the initial feature selection phase which is unsupervised. Better supervised methods may be considered to elicit a limited number of important features.
Complex concept definitions may turn out to be harder to learn, especially when nested restrictions are considered although such definitions are hardly found in KGs.
The hierarchical model in its current implementation would require a more careful search of optimal values for other hyperparameters. This also calls for experiments involving larger datasets in terms of the number of individuals.

On the ground of similar works on mixture models \cite{Ghahramani93,Tresp97mlj,Fanizzi22sitis} various extensions are possible along different lines including:
\begin{itemize}
	\item the incorporation of existing rules or axioms in the probabilistic models;
	\item a different classification setting, considering decision procedures that admit rejection cases, to better comply with the original semantics;
	\item a deeper investigation on the adoption of a Bayesian approach to adapt the structure of the models (and parameters fitting) considering suitable priors;
	\item the integration of continuous (Gaussian) features to include in the model also restrictions on numerical datatypes. 
\end{itemize}

Applications of the probabilistic classifiers to various other tasks are possible, such as KG \emph{debugging}, e.g.\ for anomaly detection, and \emph{knowledge refinement}, e.g.\ using these models for axiom discovery, etc.

\subsubsection*{Acknowledgments.}
The authors are thankful to the anonymous reviewers for their careful reading and their insightful comments and suggestions.\\
This work was partially supported by project \textit{FAIR - Future AI Research} (PE00000013), spoke 6 - \textit{Symbiotic AI}, within the NRRP MUR program funded by the NGEU package and by project \textit{HypeKG - Hybrid Prediction and Explanation with Knowledge Graphs} (H53D23003700006), under the PRIN 2022 program funded by MUR.

\bibliographystyle{splncs04}
\bibliography{biblio}

\end{document}